 \documentclass[pmlr,twocolumn]{jmlr} 

\usepackage{booktabs}
\usepackage{color}
\usepackage{graphicx} 
\usepackage{color}
\usepackage{amsmath}
\usepackage{amssymb}
\usepackage{svg}
\usepackage{caption}

\usepackage{siunitx}
\usepackage{booktabs}
\usepackage{textcomp}
\usepackage[normalem]{ulem}

\newcommand\model[1]{\texttt{#1}}

\newcommand\commentout[1]{}
 \usepackage{siunitx}

\theorembodyfont{\upshape}
\theoremheaderfont{\scshape}
\theorempostheader{:}
\theoremsep{\newline}

\jmlrvolume{}
\jmlryear{2020}
\jmlrsubmitted{}
\jmlrpublished{}
\jmlrworkshop{Machine Learning for Health (ML4H) 2020} 

\title[Short Title]{Full Title of Article\titlebreak This Title Has
A Line Break}

\title[Attend and Decode: 4D Functional MRI Task State Decoding]{Attend and Decode: 4D fMRI Task State Decoding\\Using Attention Models}

  \author{%
   \Name{Sam Nguyen} \Email{samiam@llnl.gov}\\
   \Name{Brenda Ng} \Email{ng30@llnl.gov}\\
   \Name{Alan D. Kaplan} \Email{kaplan7@llnl.gov}\\
   \Name{Priyadip Ray} \Email{ray34@llnl.gov}\\
   \addr Lawrence Livermore National Laboratory, Livermore, CA, USA
  }

\begin{document}

\maketitle

\begin{abstract}
Functional magnetic resonance imaging (fMRI) is a neuroimaging modality that captures the blood oxygen level in a subject's brain while the subject either rests or performs a variety of functional tasks under different conditions. Given fMRI data, the problem of inferring the task, known as \emph{task state decoding}, is challenging due to the high dimensionality (hundreds of million sampling points per datum) and complex spatio-temporal blood flow patterns inherent in the data. In this work, we propose to tackle the fMRI task state decoding problem by casting it as a 4D spatio-temporal classification problem. We present a novel architecture called \emph{Brain Attend and Decode} (BAnD), that uses residual convolutional neural networks for spatial feature extraction and self-attention mechanisms for temporal modeling. We achieve significant performance gain compared to previous works on a 7-task benchmark from the large-scale Human Connectome Project-Young Adult (HCP-YA) dataset. We also investigate the transferability of BAnD's extracted features on unseen HCP tasks, either by freezing the spatial feature extraction layers and retraining the temporal model, or finetuning the entire model. The pre-trained features from BAnD are useful on similar tasks while finetuning them yields competitive results on unseen tasks/conditions.
\end{abstract}

\section{Introduction}
\label{sec:intro}

Functional MRI is a neuroimaging modality that measures spatial differences in blood oxygenation over time.
As opposed to forming a static structural image of the brain, fMRI captures temporal activity patterns. 
A contrast mechanism  is used to generate the Blood Oxygen Level Dependent (BOLD) signal which may be used to infer brain activities by measuring relative oxygen concentrations in the blood \cite{Webb2003-de}. \commentout{Active brain regions elicit greater concentrations of oxygenated blood, resulting in greater contrast in the fMRI images. The ability to record whole brain activity in a noninvasive manner makes fMRI an attractive option for studying brain function.}
During the fMRI experiment, a subject lies in an MRI scanner while performing various tasks to activate different brain regions. The goal of fMRI task state decoding is to map the fMRI sequence to the task that was performed.

\commentout{The fMRI sequence can be viewed as a movie of 3D volumes (\emph{frames}), that contains complex spatio-temporal patterns.}
Functional MRI modality can help clinicians diagnose neurological diseases as well as monitor the effectiveness of therapy. For example, resting state fMRI identified significant disruptions in default mode networks coactivation in patients with Alzheimer's disease \cite{greicius2004default,rombouts2009model}. In the case of Parkinson disease, which is a neurological movement disorder, \cite{tada1998motor,sabatini2000cortical} identified lowered activation in the supplementary motor area during movement in Parkinson versus controls. These signatures can potentially be used to diagnose neurological disorders in a noninvasive manner.

Our paper's main focus is on the problem of fMRI task state decoding. Our attention-based model is geared towards the representation of complex spatio-temporal signals that fMRI data are comprised of. Beyond the formulation and validation of the models that we present in this paper, it may be possible to use them to (1) analyze the differences in fMRI data from patients versus controls, (2) directly classify neurological disorders, or (3) distinguish various pathological subtypes of a neurological disease to monitor impact of therapy specific to that subtype.

While many analytic techniques, such as \cite{naselaris2011encoding,wen2018deep}, have been developed for fMRI task state decoding, many challenges remain. Modern scanners are constantly improving its acquisition rates (number of frames per second) and its image resolution (number of pixels per frame). The data takes up significant memory and thus requires scalable solutions. Moreover, the phenomenon being measured -- the blood oxygenation levels within the brain -- is of high spatio-temporal complexity. To model this phenomenon, we need an expressive enough model that can capture this complexity. In this paper, we ``leave no stone unturned'' by using as much information as possible that is inherent in the raw fMRI data. We explicitly account for temporal dynamics in our model, and use it to compute spatio-temporal maps to indicate the relative importance of brain regions for the task decoding.

Our proposed model, called \emph{Brain Attend and Decode} (\model{BAnD}), has been validated using the Human Connectome Project (HCP) \cite{Van_Essen2013-xf,Glasser2016-kr} neuroimaging dataset, one of the most prevalent datasets for brain imaging studies. HCP contains multiple types of neuroimaging, demographic information, and cognitive scoring of over 1,000 healthy young adults (HCP-YA). Our approach operates on volumetric fMRI data and does not require any computationally expensive parcellation techniques.

\commentout{Resting state dynamics are of interest, and several connectivity patterns are known to emerge \cite{Van_den_Heuvel2010-ax}.}

The key contributions of this work are as follows:
\begin{itemize}

\item We propose the first attention-based model for processing 4D spatio-temporal data. Based on the multi-head self-attention model \cite{vaswani2017attention} and the pooling operation \cite{devlin2018bert}, we developed a self-attention model that attends to a series of embedded 3D frames.
\item Our model achieves substantial improvement in classification performance over previous works on task state decoding using the 7-task benchmark from the HCP dataset.
\item We demonstrate transfer learning of the learned features under different conditions and tasks. 
\commentout{\item We visualize the relative importance of brain regions and frames, with respect to task state decoding, by computing spatial and temporal maps.}

\end{itemize}

\section{Related Work}
Recently, there has been much interest in applying machine learning to fMRI data \cite{pereira2009machine}. Convolutional neural networks (CNNs) \cite{lecun1998gradient} have been successfully adopted to learn representational features from fMRI data. \cite{huang2017modeling} proposed an architecture based on the sparse convolutional autoencoder to learn high-level features from handcrafted time series derived from the raw fMRI data. \cite{wang2018task} proposed a 4-layer CNN that classifies tasks from the raw fMRI voxel values. Both works were demonstrated on HCP, treating the entire fMRI sequence as input.

More recent efforts explore the use of sequence models, that ingest each frame of the fMRI sequence in a sequential order. As such, sequence models are designed to capture temporal correlations between consecutive frames. \cite{mao2019spatio} used a CNN to extract features from each fMRI frame, then applied each frame's feature as input into a Long Short-Term Memory (LSTM) \cite{hochreiter1997long} model. \commentout{An LSTM is a type of recurrent neural network (RNN) that takes in sequence inputs, generally in forward order where input from $t-1$ is ingested before input from $t$. It is designed to retain long-term dependencies within the sequence, by the use of gates that allow signals to bypass so it can retain information across longer time steps.} Furthermore, combinations of LSTM and CNN have become a popular approach for fMRI analysis. \cite{thomasfmri} used a bidirectional LSTM \cite{birnn} in conjunction with a simple CNN. \commentout{(A bidirectional LSTM is a pair of LSTMs, in which one LSTM processes the data in forward order and another LSTM processes the data in backward order, then the result is merged.)} \cite{voxelwise3d2018} used a LSTM in conjunction a residual network (ResNet) \cite{he2016deep}.

\vspace{-1.5em}
\section{Methods}
\label{sec:methods}
Unlike RNNs and LSTMs that process data sequentially, transformers \cite{vaswani2017attention} offer a different way to reason about sequence data, by \emph{not} treating them as sequence, but rather, processing the entire sequence of data all at once. This allows frames that are temporally far apart to have the same level of access to each other's information, in stark contrast to RNNs and LSTMs. This non-sequential access allows transformers to outperform RNNs in terms of long-term memory, as well as scalability due to its parallelizable architecture. \commentout{In addition, the attention scores computed as part the transformer computations can be used to trace the flow of information \cite{devlin2018bert,videotransformernet} within the model.} Due to these advantages, we propose the first 
transformer-based model for 4D fMRI data, and demonstrate its utility in fMRI task state decoding.
\vspace{-0.8cm}

\subsection{Problem Statement}

An instance of the fMRI data is a tuple $(\mathbf{x}_{1:T}, \mathbf{y})$, where $\mathbf{x}_{1:T} = \{\mathbf{x}_1, \mathbf{x}_2, ..., \mathbf{x}_T\}$ is a sequence of $T$ 3-D frames. Each frame $\mathbf{x}_t \in \mathbb{R}^3$ is a snapshot of the subject's brain activity at time $t$. The tasks and conditions under which $\mathbf{x}_{1:T}$ was captured during the fMRI procedure are denoted as $\mathbf{y}$. 

\subsection{Frame Embedding}
First, we need a model to transform each fMRI frame into a compressed numeric representation, or \emph{embedding}. 
\subsubsection{3D CNN}
Using 3D CNN in fMRI task state decoding was proposed by \cite{wang2018task}. The 3D kernels in the 4 convolutional layers are 1x1x1, 7x7x7, 3x3x3 and 3x3x3, respectively. Except for the first layer, whose stride is 1, all other layers adopt a stride of 2. This model uses global average pooling \cite{he2016deep} 
instead of max pooling on the final feature maps. The resulting feature maps are flattened (with dimension 512, which is the number of output channels in the CNN) and fed into a classifier with 2 fully-connected layers, whose output dimensions are 64 and 7, respectively. This model accounts for the temporal dimension of fMRI data by using 1x1x1 convolution (thus treating the temporal dimension as extra channels) to weigh every voxel accordingly across the time series. This model is implemented as a baseline method and will be referred to as \model{CNN}.

\vspace{-1em}
\subsubsection{3D ResNet}
The 3D ResNet model \cite{voxelwise3d2018} extends \model{3DCNN} by adopting residual connections and bottleneck blocks from ResNet. Since no code was made available, we implemented the 3D ResNet model based on our interpretation of the model description given in  \cite{voxelwise3d2018} as follows: We augmented the 2D ResNet-18 \cite{he2016deep} into its 3D counterpart, the 3D ResNet-18 (cf. Table \ref{tab:3dresnet18}). This model is implemented as a baseline method and will be referred to as \model{3DResNet}. For our proposed model, we make use of the spatial extractor, 3D ResNet-18, within the \model{3DResNet}.\commentout{, because it extracts superior features than the 3D CNN within the \model{3DCNN}.}

\subsection{Temporal Modeling}

\subsubsection{LSTM-based Models}
In \cite{mao2019spatio} and \cite{voxelwise3d2018}, an LSTM was used to model the temporal correlation between the embedded frames. The LSTM model has 2 hidden layers, each with 128 hidden units. In \cite{thomasfmri}, a bidirectional LSTM was used instead. This \model{3DResNet-LSTM} model -- the combination of 3D ResNet and LSTM -- is implemented as a baseline method.

\subsubsection{Self-attention and Transformers}

We hypothesize that attention is valuable in modeling the temporal dynamics of fMRI data. The transformer \cite{vaswani2017attention} applies self-attention to the entire sequence to model relationships between the frames in the sequence. These relationships are quantified as attention scores, which weight each frame due to its relevance to other frames. Through empirical experiments, we optimized the number of attention layers to 2 and attention heads to 8 for the transformer in our proposed model. 

\subsection{Our Transformer-based Model}

\begin{figure*}[t]
  \centering
  
  \includegraphics[width=0.8\textwidth]{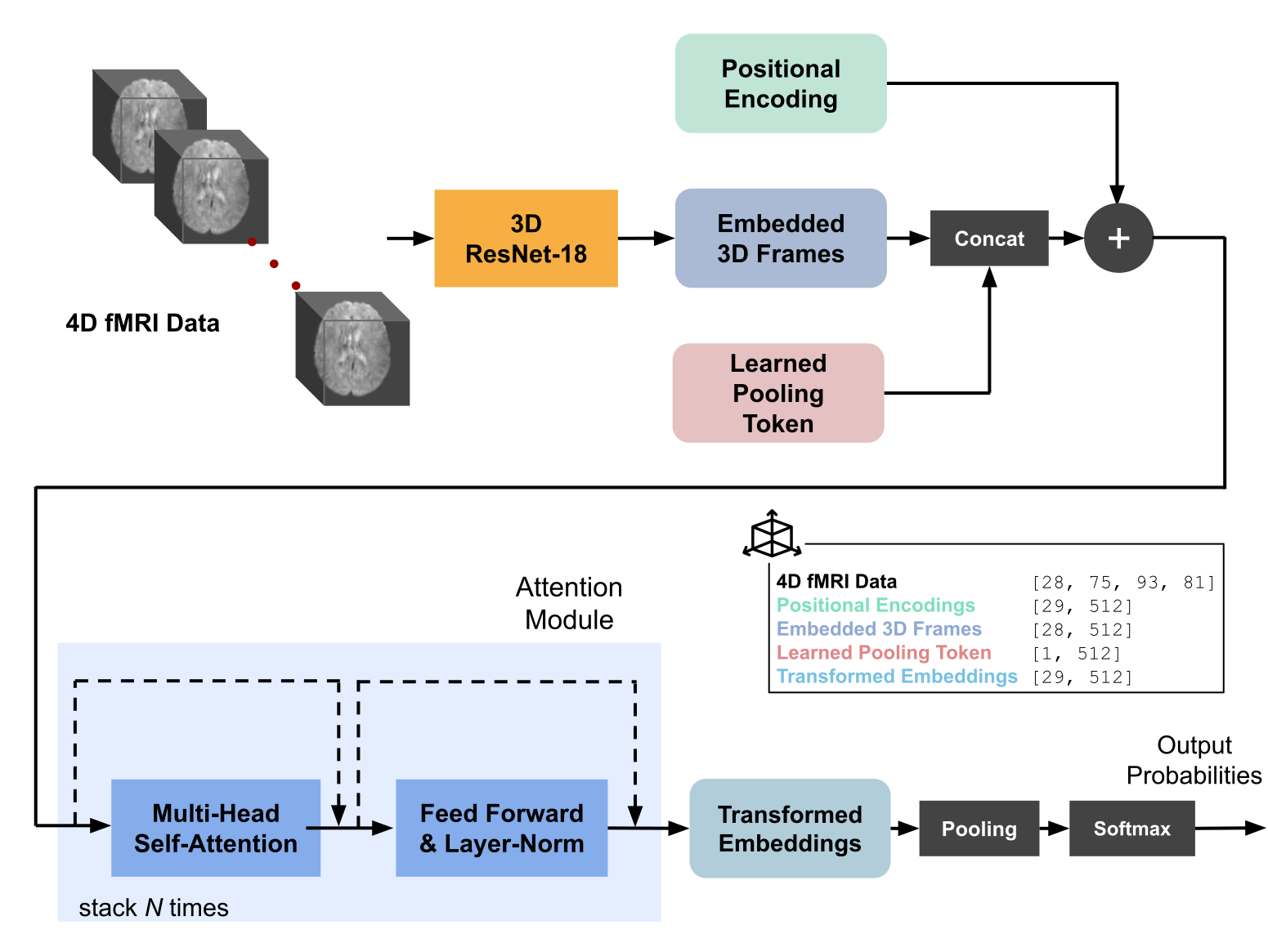}
  \caption{BAnD architecture. Our model has $N=2$. We show input dimensions at each component for the HCP 7-task dataset.}
  \label{fig:band}
\vspace{-1em}
\end{figure*}

Our model, called \emph{Brain Attend and Decode} (\model{BAnD}) (cf. Figure \ref{fig:band}), adapts the transformer model to the task of fMRI task state decoding. The transformer maps each embedded 3D frame $\mathbf{e}_t = \sigma(\mathbf{x}_t)$ into a latent vector $\mathbf{z}_t$ that is conditioned with information from other embedded frames. To combine these latent vectors (also known as \emph{transformed embeddings}) into a single representation that can be fed into the classifier , we adopt the approach in \cite{devlin2018bert}, whereby we insert a shared artificial token ($\mathbf{e}_\star$) to the beginning of each series and train it together with the transformer. This can be understood as the one ``frame'' that extracts essential information from the entire time series. By visualizing the attention of that frame on the other frames, this provides insights about which frames are important to the classification problem. The workflow of \model{BAnD} is summarized in Figure \ref{fig:band}.

Similar to 3DResNet-LSTM, we apply the base 3D ResNet-18 to embed each frame of the data, i.e., $\mathbf{e}_t = \sigma(\mathbf{x}_t)$ for $t=1,...,T$.  A special pooling token $\mathbf{e}_\star$ is concatenated to this series, i.e., $\{\mathbf{e}_\star, \mathbf{e}_1, ..., \mathbf{e}_T\}$.  Positional encoding \cite{vaswani2017attention} is added to each embedding to enable position understanding.  The transformer will map these embeddings $\{\mathbf{e}_\star, \mathbf{e}_1, ..., \mathbf{e}_T\}$ into latent vectors $\{\mathbf{z}_\star, \mathbf{z}_1, ..., \mathbf{z}_T\}$.  The pooling token $\mathbf{z}_\star$ is fed into a classifier (comprised of 2 fully-connected layers with output dimensions of 512 and the number of classes) to get the logits for task classification. Finally, softmax is applied to get output probabilities for each class.
We applied dropout with a rate of 0.2 to the transformer and classifier layers. 

\begin{table}[]
\tiny
\begin{tabular}{@{}cc@{}}
\toprule
\textbf{Layer Name}                & \begin{tabular}[c]{@{}c@{}} \textbf{3D ResNet-18}\\ (kernel size, number of filters, stride)\end{tabular} \\ \midrule
conv1      &  \multicolumn{1}{c}{$7 \times 7 \times 7, 64$, stride $2$}                                                                          \\ \midrule
conv2      &  \multicolumn{1}{c}{$\left[ \begin{array}{cc} 3 \times 3 \times 3, 64, 1 \\ 3 \times 3 \times 3, 64, 1 \end{array}\right] \times 2$} \\ \midrule 
conv3      &  \multicolumn{1}{c}{$\left[ \begin{array}{cc} 3 \times 3 \times 3, 128, 2 \\ 3 \times 3 \times 3, 128, 2 \end{array}\right] \times 2$} \\ \midrule 
conv4      &  \multicolumn{1}{c}{$\left[ \begin{array}{cc} 3 \times 3 \times 3, 256, 2 \\ 3 \times 3 \times 3, 256, 2 \end{array}\right] \times 2$} \\ \midrule 
conv5      &  \multicolumn{1}{c}{$\left[ \begin{array}{cc} 3 \times 3 \times 3, 512, 2 \\ 3 \times 3 \times 3, 512, 2 \end{array}\right] \times 2$} \\ \midrule 
global average pool &  \multicolumn{1}{c}{$1 \times 1 \times 1 \times 512$} \\ \midrule
flatten &  \multicolumn{1}{c}{embedding of size $512$}
\\ \bottomrule
\end{tabular}
\caption{Base 3D ResNet-18 architecture}
\label{tab:3dresnet18}
\vspace{-1em}
\end{table}

For BAnD and 3DResNet-LSTM, because we embedded each 3-D frame individually before passing the embeddings to the temporal model, limited GPU memory was an issue. By employing model parallelization, we were able to increase our batch size from 2 to 64, by training across 8 nodes utilizing a total of 32 GPUs during training.

\vspace{-0.5cm}
\section{Datasets}


\subsection{HCP 7-task}
The HCP-YA fMRI dataset includes 7 functional tasks, each under different conditions. These 7 tasks as a whole provide good brain activation coverage \cite{Barch2013-wy}, thus a classifier trained on this dataset would be useful for brain state decoding on a wide range of functional tasks. We used the same data as \cite{wang2018task} (cf. Table \ref{tab:7task}), which is a subset of the full HCP-YA fMRI data. Like \cite{wang2018task}, we cropped out empty regions in the raw fMRI data, so the spatial dimensions $[d_x, d_y, d_z]$ are reduced from $[91, 109, 91]$ to $[75, 93, 81]$. 
The data from \cite{wang2018task} included 8 extra seconds post task/condition. We omitted these extra post frames (sampled at $0.72$ frames per second) and examined only the frames during the actual task/condition. Depending on the task $c$, our data instance varies in its dimensions $[T_c, d_x, d_y, d_z]$ where $[d_x, d_y, d_z]$ are same as above, but $T_c$ ranges from $16$ to $38$.

Only 965 of the 1,034 subjects performed all 7 tasks in this benchmark. For each task, a subject might have performed more than one run. The total number of 4D data instances in this dataset is 17,368, with class distribution as shown in Table \ref{tab:7task}. The 17,368 instances were partitioned into train set (70\%), validation set (10\%) and test set (20\%). Because each subject might perform a task multiple times, to prevent leakage of information across train/validation/test partitions, we stratified the splits so that data from a subject only belongs to one of the partitions.
\vspace{-0.5cm}

\begin{table*}[]
\small
\centering
\begin{tabular}{@{}llrrr@{}}
\toprule
Task - Condition & \# Frames & \# Instances & Percentage \\ \midrule
Emotion - Fear & 25 & 2,895 & 16.7\% \\
Gambling - Loss & 38 & 1,930 & 11.1\% \\
Language - Present story & 28 & 3,860 & 22.2\% \\
Motor - Right hand & 16 & 1,930 & 11.1\% \\
Relational - Relation & 22 & 2,895 & 16.7\% \\
Social - Mental & 31 & 2,893 & 16.6\% \\
WM - 2-back places & 38 & 965 & 5.6\% \\ \bottomrule
\end{tabular}
\caption{HCP 7-task dataset summary. Each instance is a complete 4D data point of a task/condition. A subject might perform a task multiple times, resulting in multiple instances. \# Frames is the time dimension of each task/condition.}
\label{tab:7task}
\vspace{-2em}
\end{table*}

\subsection{Transfer learning}
We distilled 3 subsets from HCP to evaluate transfer learning under intra-task (same task, different conditions) and inter-task (different tasks) conditions. 

\textbf{HCP 7-task-b set}: This dataset was extracted from HCP and has similar characteristics as the HCP 7-task subset that we used to train BAnD. This dataset is used for the intra-task transfer evaluation. The fMRI tasks are the same, but the conditions are different. Instead of the conditions in Table \ref{tab:7task}, we have: neutral, win, math, right foot, match, random, 2-back body, for the respective tasks. Pre-processing and augmentation scheme remain the same. There are 18,896 data instances in total.

\textbf{Motor transfer set}: In this setting, BAnD was trained with data instances from the following 6 tasks: Working memory, Gambling, Relational, Social, Language, Emotion. Then the model was evaluated on the held-out Motor task, under 5 different conditions: right hand, left hand, right foot, left foot and tongue. There are 9,650 data instances in total.

\textbf{WM transfer set}: This dataset is similar to the Motor transfer set, except the Working Memory (WM) task is held out instead of Motor.  There are 3,860 data instances in total.
\vspace{-1em}

\subsection{Data Processing}
It was important to normalize each data instance to have the same number of timesteps. Otherwise, the temporal model would simply learn to count frames to infer the task/condition. 
Following the data processing scheme in \cite{wang2018task}, we extracted sets of $k$ contiguous frames from each data instance (i.e., each set is $\mathcal{S} = \{\mathbf{x}_r, \mathbf{x}_{r+1}, ..., \mathbf{x}_{r+k-1}\}$ where $r$ is a random index) for training. During validation and testing, only the set corresponding to $r=0$ (the first $k$ frames) were selected. For data instances with less than $k$ frames, we looped the time series until $k$ frames are achieved, as is commonly done in video recognition \cite{carreira2017quo}. In our experiments, $k = 28$.

\vspace{-1.5em}
\section{Results and Analysis}

In addition to the deep-learning-based baseline models described in section \ref{sec:methods}, we also compared against other traditional baselines that use parcellated fMRI data. Due to the lack of code and data processing scripts from previous works, we implemented all baseline models from scratch, with the intent to share our code to promote reproducibility within the research community. For linear models, we used Scikit-learn package\footnote{https://scikit-learn.org/, version 0.22.1}. For neural network models, we used PyTorch\footnote{https://pytorch.org/, version 1.2} and adapted the transformer code from the PyTorch-Transformers repo\footnote{https://github.com/huggingface/pytorch-transformers}.

For all our experiments, we used an IBM Power8 computing cluster, where each node is equipped with 4 Nvidia P100 GPU cards.

\subsection{HCP 7-task}

\subsubsection{Comparison with Methods Using Parcellated fMRI Data}
We validated our model against a regression-based baseline that utilizes parcellated fMRI data. We used the parcellated data that came with the HCP-YA dataset, which had undergone the Glasser parcellation \cite{Glasser2016-qg}. Each data instance contains 360 cortical and 14 subcortical regions.
Prior to computing the correlation matrix, we applied global signal regression to subtract out common trends across the timeseries.
This is performed using a 2 parameter linear regression capturing shift and scale perturbations from the global average.
We then constructed a connectome matrix from the fMRI data \cite{Behrens2012-vy}.
We next utilized the Pearson Correlation matrix between parcellated regions to represent the amount of functional connectivity between regions \cite{Sporns2011-zc}.
Then, we band-filtered the timeseries signals to the range 0.009 Hz - 0.25 Hz using a 4-th order Butterworth filter. The correlation step results in a $374 \times 374$ matrix with values ranging from -1 to 1.

We performed this regression task in two ways: logistic regression and fully connected neural network. For logistic regression, we experiment with non-regularized model as well as an L1 regularized version with k-fold cross-validation. We set k to be 3 and tuned the inverse of regularization strength in the range of $10^c$ with $c$ in $[-3, -2, .., 3]$. As for the neural network model, we build a 3-layer fully-connected network with ReLU activation. Its output is fed into a softmax layer and and the model is trained with cross-entropy loss. We did light hyper-parameter tuning and the size of the hidden layer was set to $320$.

\subsubsection{Comparison with Methods Using Non-Parcellated fMRI Data}

Results for each task are reported with means and standard deviations from 3 different random splits of training, validation and testing data, whenever applicable. For neural network models, we used Adam optimizer \cite{kingma2014adam} with weight decay of $0.0001$. We empirically finetuned the learning rates for each model and used a cosine annealing learning rate schedule.

BAnD outperforms all other models, as shown by the increasing accuracy values in Figure \ref{tab:classification}: 
 from 91.4\% with 3DCNN, to 94.5\% with 3DResNet, to 95.1\% with BAnD. We conjecture that this increase in performance stems from the added attention capability to understand complex spatio-temporal dynamics in 4D fMRI data. It is also important to note that with a logistic regression model, correlation matrix regression approach with L1 regularization achieved an accuracy of 92.2\%, which is better than 91.4\% of the 3DCNN model. However, extensive pre-processing is required to parcellate raw fMRI data and compute correlation matrices. Performance at down-stream tasks might also depend on different parcellation strategies. It is further unclear how to extract/analyze the temporal dynamics inherent in fMRI data with this approach.

As for the 3DResNet-LSTM model, even though we experimented with an extensive hyperparameter sweep of the number of hidden layers $[1, 2, 3]$, the number of hidden units for each layer $[64, 128, 256]$, the directionality (bi- or uni-) of the LSTM model, as well as different dropout rates $[0.2, 0.4, 0.6]$, the model did not converge to a satisfactory validation/test accuracy. With limited computational budget and time, we trained each of these configurations for 24 hours under a distributed environment with 32 GPU cards. Further work is needed to analyze why these LSTM models did not converge under this HCP benchmark. Nonetheless, our BAnD model was able to converge in under 12 hours with the same computational setting. This might be due to the Transformer's advantage of parallelizability and faster convergence, as suggested by \cite{vaswani2017attention}.

Furthermore, as \cite{popel2018training} noted, a bigger batch size (up to a certain threshold) yields better results with the Transformer model. Due to limited GPU memory, BAnD was trained with a batch size of 2 on each GPU card. To increase the batch size, we froze the 3D ResNet-18 layers in the converged BAnD model and trained a new Transformer on top. With this trick, we were able to increase the batch size from 2 to 128, thus deriving at a new model called BAnD++. From Table \ref{tab:classification}, BAnD++ shows a significant increase in classification accuracy, from 95.1\% to 97.2\%. 

\begin{figure}[t]
\centering
\includegraphics[width=0.5\textwidth]{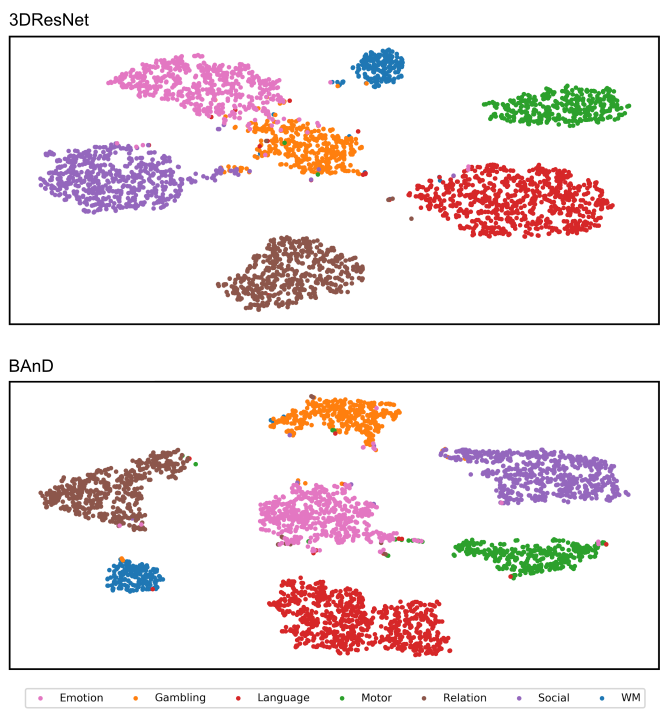} 
\caption{3DResNet and BAnD embeddings t-SNE. For 3DResNet, which does not model temporality explicitly, the clusters are more tangled, while for BAnD, which learns the temporal dynamics in 4D fMRI through Transformer, the clusters are more separated, resulting in better classification performance.}
\label{fig:tsne}
\end{figure}

To evaluate the results qualitatively, we applied t-SNE \cite{maaten2008visualizing} to visualize the final embeddings generated by the 3D ResNet and BAnD++ models.
As can be seen in Figure \ref{fig:tsne}, 3DResNet, though quite effective, was not able to disambiguate clearly the data instances belonging to Emotion, Gambling and Social tasks. On the other hand, BAnD++, with the added capability of a Transformer, was able to separate out those instances into separate groups.

Interestingly, even though the joint model 3DResNet-LSTM did not converge to satisfactory classification performance in our experiments, a model with a 3D ResNet-18 spatial model pre-trained as BAnD, with an LSTM temporal head (denote this model as 3DResNet-LSTM++) achieves similar results to BAnD++. This shows that BAnD with the Transformer head instead of an LSTM head not only made it possible for the spatial feature extractor to converge faster but the learned features are also general enough for a new LSTM to learn the temporal patterns of the 4D fMRI data series, and achieves significant performance.

\begin{table}[]
\small
\centering
\begin{tabular}{@{}ll@{}}
\toprule
Logistic regression & $0.915  \pm 0.003 $ \\
Logistic regression (L1 reg) & $0.922 \pm 0.002 $ \\
Fully-connected NN & $0.915 \pm 0.004 $ \\
\midrule
3DCNN & $0.914 \pm 0.0204 $ \\
3DResNet &  $0.948 \pm 0.011$ \\ 
3DResNet-LSTM &  $0.534 \pm 0.053$ \\ 
BAnD (ours) & $0.951 \pm 0.0062$ \\ \midrule 
BAnD++ (ours) &  $\textbf{0.972} \pm \textbf{0.0057}$ \\ 
3DResNet-LSTM++ & $0.970 \pm 0.0037$ \\ \bottomrule
\end{tabular}
\caption{Accuracy results for different models on the HCP 7-task. Means and standard deviations are calculated from 3 different splits of training, validation and testing data.} 
\label{tab:classification}
\end{table}

\begin{figure*}[t]
\centering
\includegraphics[width=0.9\textwidth]{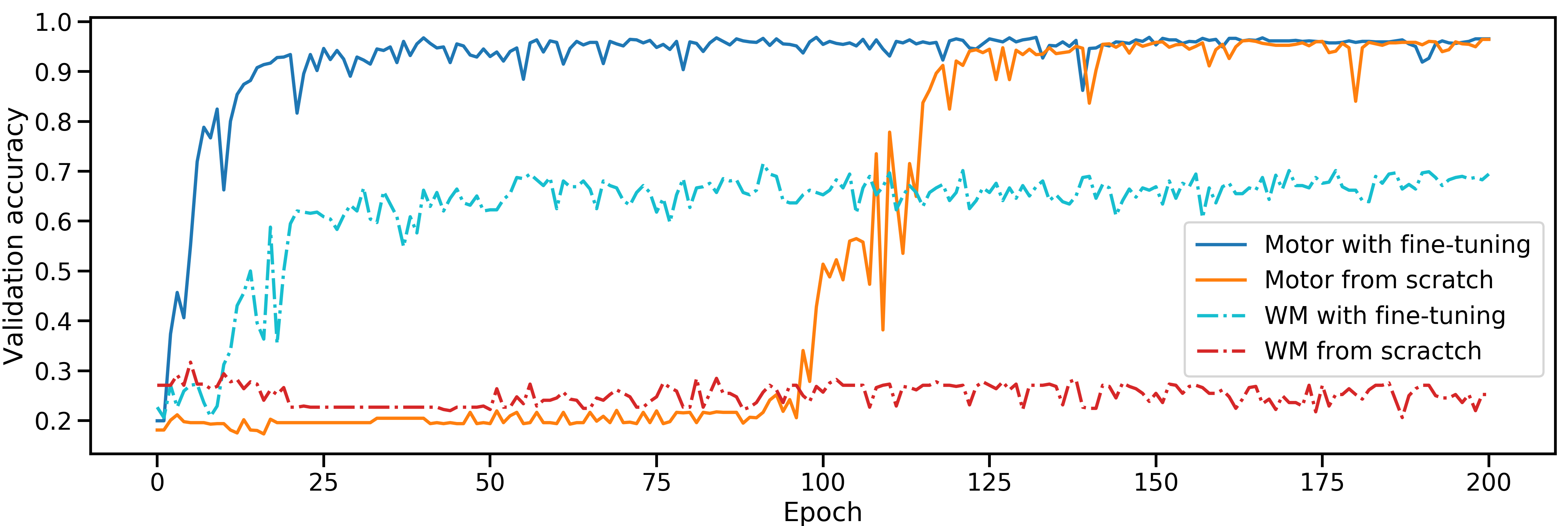}
\caption{Finetuning vs training from scratch}
\label{fig:transfer}
\end{figure*}

\vspace{-1.0em}
\subsection{Transfer learning}
In this section, we discuss the experimental results on the transferability of BAnD features under the intra-task and inter-task settings, as described above.

\textbf{Intra-task, HCP 7-task-b set:} The goal of this subset is to investigate whether the features learned with BAnD can transfer to a similar dataset. We first froze the spatial feature extractor (i.e., the 3D ResNet-18 layers) in BAnD trained on HCP 7-task dataset, added a new temporal Transformer head (with the same hyperparameters) and trained it on HCP 7-task-b dataset, without finetuning the 3D ResNet-18 layers. As can be seen in Table \ref{table:transfer}, our model achieved 93.6\% accuracy in this setting, which suggests that the spatial features learned with BAnD are highly applicable to similar tasks.

\textbf{Inter-task, Motor transfer:} Under this setting, the test tasks are significantly different from the transfer tasks. With the same approach as applied to the HCP 7-task-b, where we froze the 3D ResNet-18 layers in BAnD and trained a new Transformer head, we observed an accuracy of 51.1\%. This result suggested that there was room for improvement. So we further finetuned the 3D ResNet-18 layers to the task of classifying the different Motor conditions. As shown in \ref{table:transfer}, we observed a significant increase in classification performance, from 51.1\% to 94.1\%. On the other hand, a model trained from scratch achieved 93.4\% classification accuracy, but took significantly more time, while the \emph{Motor with finetuning} model was able to leverage features learned from the tasks it was pre-trained on.

\begin{table}[]
\centering
\begin{tabular}{@{}ll@{}}
\toprule
7-task-b & $0.936 \pm 0.012$ \\
Motor tasks &  $0.511 \pm 0.027 $ \\ 
WM tasks &  $0.623 \pm 0.031 $ \\  \midrule
Motor tasks with finetuning &  ${0.941} \pm 0.0014$ \\ 
Motor tasks from scratch &  ${0.934} \pm 0.0067$ \\ 
WM tasks with finetuning &  ${0.71 \pm 0.0043}$ \\
WM tasks from scratch &  $0.23 \pm 0.018$ \\ \bottomrule
\end{tabular}
\caption{Transfer learning and finetuning accuracy results}\smallskip
\label{table:transfer}
\vspace{-1em}
\end{table}

\textbf{Inter-task, Working Memory transfer:} We repeat the experiment above with the held-out WM tasks instead of Motor tasks. We noticed the same patterns as observed in the unseen Motor tasks, in which further finetuning helped our model achieve a significant increase in performance and with shorter training time. 

\begin{figure}[t]
\label{fig:resnet_tsne}
\centering
\includegraphics[width=0.5\textwidth]{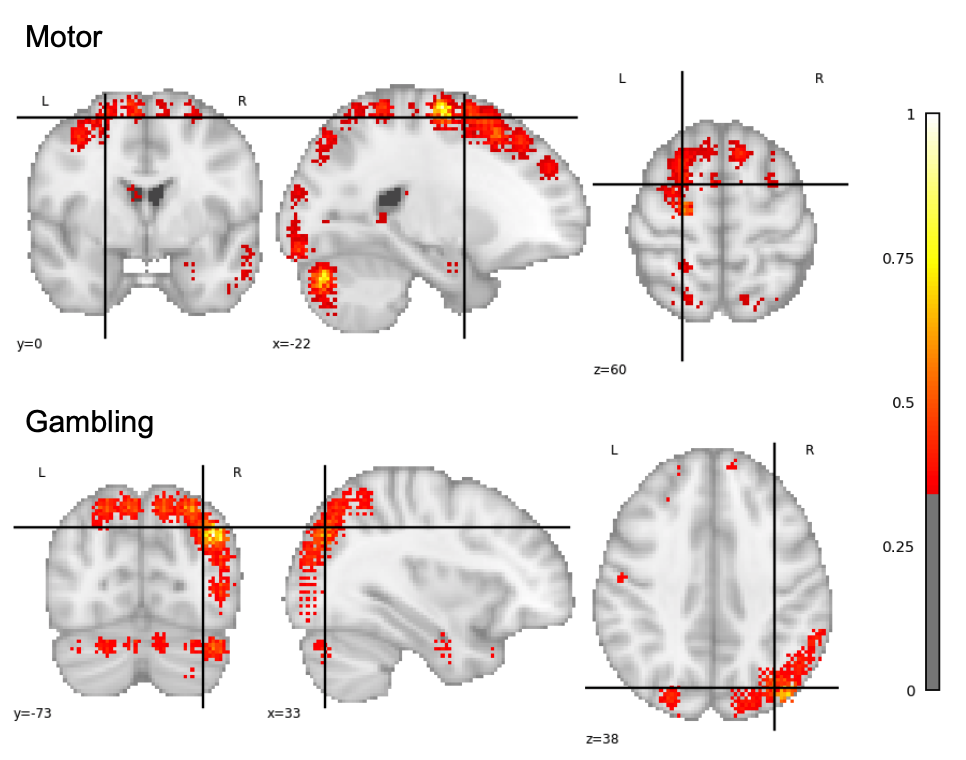} 
\caption{BAnD spatial activation map for Motor (Right hand) and Gambling (Loss) task}
\label{fig:spatial_map_motor}
\vspace{-0.5em}
\end{figure}

\vspace{-0.2cm}
\subsection{Analysis}
\begin{figure}[t]
\centering
\includegraphics[width=0.5\textwidth]{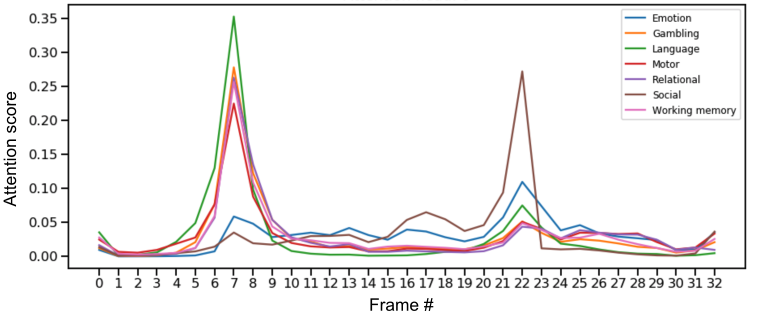}
\caption{BAnD temporal activation map}
\label{fig:temporal_map}
\vspace{-1em}
\end{figure}

\textbf{Temporal activation maps:}\space\space\space
Attention models such as Transformer allows one to examine the attention weights of the model to see which frame is weighted more for a classification result. However, our base BAnD model was trained with 8 attention heads, which makes it hard to discern which set of attention weights is truly important for a classification. Thus, we trained a new BAnD++ with only 1 attention head to make it easier to visualize. To show attention across more frames, we visualize the model's attention weights across $k=32$ frames. We calculated the attention weights across time for each data instance from the test set that was correctly classified by the model. We then averaged all attention weights in a task to get a representative set of weights. Figure \ref{fig:temporal_map} shows the result for the 7 tasks in HCP 7-task benchmark
Interestingly, across different tasks, the 7th frame of a data series seems to have a strong influence on the predictions. This can be due to a delay in the BOLD signal in response to a stimuli \cite{liao2002estimating}.

\textbf{Spatial activation maps:}\space\space\space
Using Grad-Cam \cite{selvaraju2017grad}, we were able to find the set of voxels that were highly activated for a certain class prediction by our BAnD model. We calculated this spatial activation map for each data series, then averaged across data points for a class and picked the 7th frame in each series because our temporal analysis showed that 7th frame was deemed important by the model.
We used the nilearn\footnote{https://nilearn.github.io/} package to project from 3D voxel space to brain statistical maps. Figure \ref{fig:spatial_map_motor} shows those for the Motor and Gambling task.

To motivate reproducible research in the area of fMRI task state decoding, we plan to release our data processing scripts, codes and pretrained models upon acceptance. 

\section{Conclusion}
In this work, we presented a novel attention-based model for processing 4D fMRI data, and showed that our proposed novel architecture, BAnD, which is based on a multi-headed self-attention module, achieves significant performance gains compared to previous works on a recent 7-task fMRI benchmark from the Human Connectome Project dataset. We further demonstrated transfer learning capability of BAnD's learned features to unseen conditions and tasks and show that BAnD achieves competitive results with finetuning. To try to understand what BAnD was attending to for its classifications, we computed both spatial and temporal activation maps, highlighting brain regions and frame important for task decoding. Future work includes transferring BAnD to other fMRI data sets, such as ADHD, or to other modalities of medical imaging.

\acks{This work was performed under the auspices of the U.S. Department of Energy by Lawrence Livermore National Laboratory under Contract DE-AC52-07NA27344 and was supported by the LLNL LDRD Program under Project No. 19-ERD-009. LLNL-CONF-788817.}

\bibliography{jmlr-sample}

\end{document}